\documentclass[11pt]{article}

\usepackage[margin=1in]{geometry}
\usepackage{amsmath,amssymb}
\usepackage{booktabs}
\usepackage{algorithm2e}

\usepackage{xcolor}
\usepackage{graphicx}
\usepackage{multirow}
\usepackage{subcaption} 
\usepackage{hyperref}

\hypersetup{colorlinks=true, linkcolor=blue, citecolor=blue, urlcolor=blue}

\title{SoftDTW-CUDA-Torch: Memory-Efficient GPU-Accelerated\\Soft Dynamic Time Warping for PyTorch}
\author{Ron Shapira Weber \and   Oren Freifeld
\and 
The Faculty of Computer and Information Science\\
Ben-Gurion University of the Negev\\
}
\date{}

\begin{document}
\maketitle

\begin{abstract}
We present \texttt{softdtw-cuda-torch}, an open-source PyTorch library for computing Soft Dynamic Time Warping (SoftDTW) on GPUs.
Our implementation addresses three key limitations of existing GPU implementations of SoftDTW: a hard sequence-length cap of 1024, numerical instability in the backward pass for small smoothing parameters, and excessive GPU memory consumption from materializing pairwise distance tensors.
We introduce (1)~tiled anti-diagonal kernel execution that removes the sequence-length constraint, (2)~a log-space backward pass that prevents floating-point overflow, and (3)~a fused distance-computation mode that eliminates the $\mathcal{O}(BNM)$ intermediate distance tensor, achieving up to 98\% memory reduction compared to prior work.
The library supports arbitrary sequence lengths, full PyTorch autograd integration, and Soft-DTW Barycenter computation.
Code is available at \url{https://github.com/BGU-CS-VIL/sdtw-cuda-torch}.
\end{abstract}

\section{Introduction}
\label{sec:intro}

Dynamic Time Warping (DTW)~\cite{sakoe1978dynamic} is a classical algorithm for measuring similarity between temporal sequences under nonlinear alignment.
While widely used in time-series classification, clustering, and retrieval, DTW is non-differentiable and therefore cannot serve directly as a training loss for neural networks.

Cuturi and Blondel~\cite{cuturi2017soft} introduced \emph{Soft-DTW}, a differentiable relaxation that replaces the hard minimum in the DTW recurrence with a smooth minimum parameterized by a temperature $\gamma > 0$.
As $\gamma \to 0$, Soft-DTW recovers classical DTW; for $\gamma > 0$ it provides a smooth, differentiable loss suitable for gradient-based optimization.
Soft-DTW has since been adopted for time-series forecasting, sequence-to-sequence training, metric learning, and differentiable alignment research~\cite{blondel2021differentiable}.

Maghoumi~\cite{maghoumi2020softdtw} provided the first open-source CUDA implementation of SoftDTW for PyTorch, enabling GPU-accelerated computation.
However, this implementation has three practical limitations that restrict its applicability: (1)~sequences are limited to length 1024 due to CUDA thread-per-block constraints, (2)~the backward pass operates in linear space and is prone to numerical overflow for small~$\gamma$, and (3)~the full pairwise distance matrix of shape $(B, N, M)$ must be materialized in GPU memory, which becomes prohibitive for long or high-dimensional sequences.

These limitations directly impeded our research on temporal alignment problems. Concretely, in our prior works on Diffeomorphic Temporal Alignment Networks (DTAN)~\cite{shapira2019diffeomorphic}, Regularization-Free DTAN~\cite{shapira2023regularization}, TimePoint~\cite{shapira2025timepoint}, and Synchronization of Multiple Videos~\cite{naaman2025synchronization}, we used SoftDTW primarily to compare it against our own methods on various benchmarks. However, even though our methods were more scalable, direct comparisons had to be limited, as SoftDTW’s GPU memory constraints forced us to either restrict sequence lengths, reduce batch sizes, or fall back to CPU implementations, severely hindering speed and scalability.

In this work, we present \texttt{softdtw-cuda-torch}, a reimplementation that addresses all three limitations while maintaining full compatibility with PyTorch's autograd system.

\begin{figure}[t]
  \centering
  \includegraphics[width=0.9\linewidth]{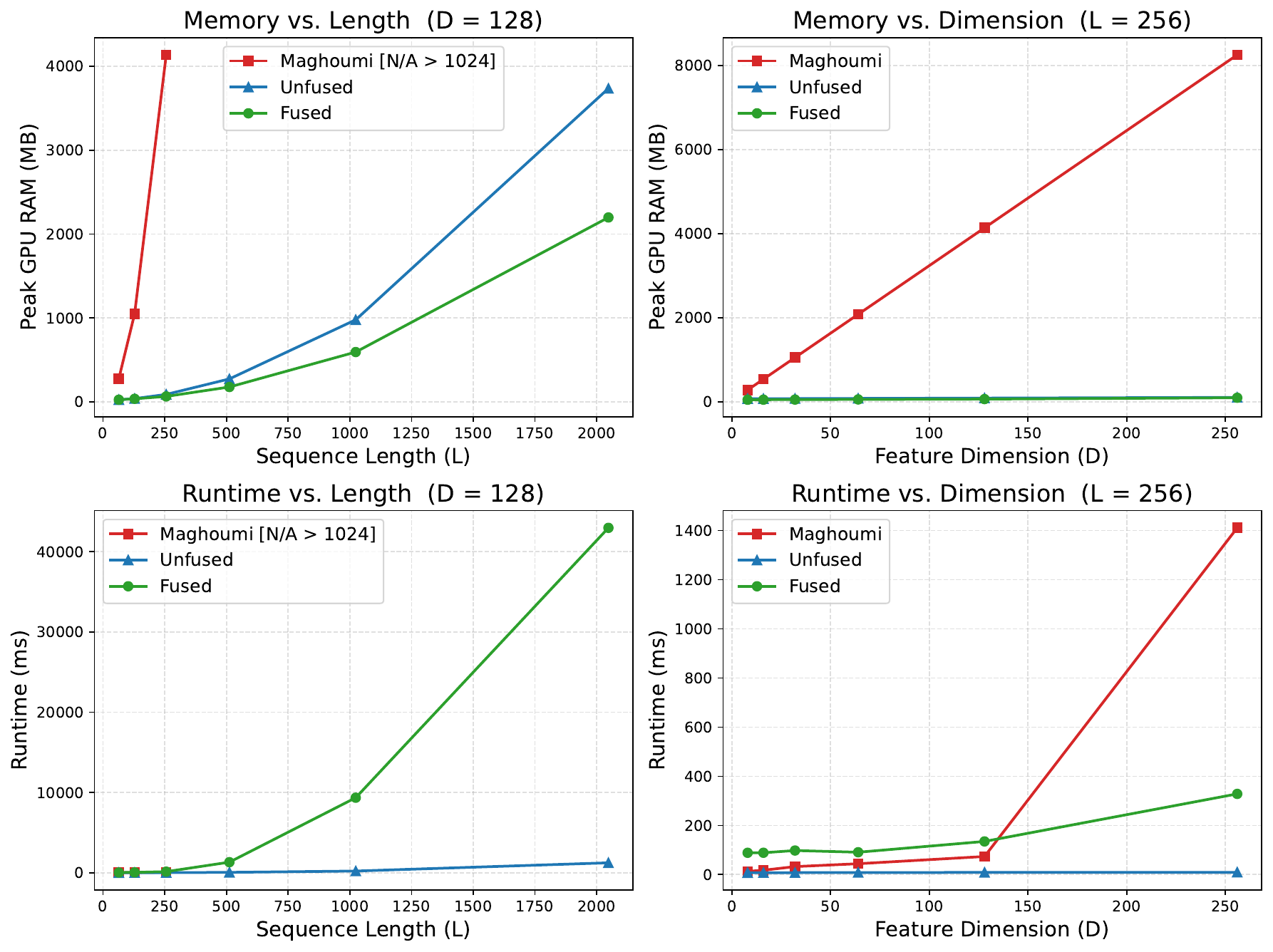}
  \caption{Benchmark results for batch size $B = 32$. \textbf{Top row:} Peak GPU memory (MB) as a function of sequence length $L$ (left, $D = 128$) and feature dimension $D$ (right, $L = 256$). \textbf{Bottom row:} Wall-clock runtime (ms) for the corresponding configurations. Maghoumi's implementation is unavailable for $L > 1024$ (CUDA thread-block limit) and runs out of memory for large configurations; our unfused and fused modes remain operational throughout. The fused mode trades runtime for significant memory savings.}
  \label{fig:benchmarks}
\end{figure}
\section{Background}
\label{sec:background}
This section describes
the original SoftDTW, based on~\cite{cuturi2017soft}. 
\paragraph{Soft-DTW formulation.}
Given two sequences $\mathbf{x} = (x_1, \ldots, x_N)$ and $\mathbf{y} = (y_1, \ldots, y_M)$ with $x_i, y_j \in \mathbb{R}^D$, and a pointwise cost $d(x_i, y_j)$ (typically the squared Euclidean distance), Soft-DTW is defined via the recurrence
\begin{align}
  R_{i,j} = d(x_i, y_j) + \mathrm{softmin}_\gamma\!\big(R_{i-1,j-1},\; R_{i-1,j},\; R_{i,j-1}\big)
  \label{eq:forward}
\end{align}
where the soft-minimum is
\begin{align}  \mathrm{softmin}_\gamma(a, b, c) = -\gamma \log\!\Big(e^{-a/\gamma} + e^{-b/\gamma} + e^{-c/\gamma}\Big)
  \label{eq:softmin}
\end{align}
with boundary conditions $R_{0,0} = 0$, $R_{i,0} = +\infty$ for $i \geq 1$, and $R_{0,j} = +\infty$ for $j \geq 1$.
The Soft-DTW distance is
\begin{align}\mathrm{sdtw}_\gamma(\mathbf{x}, \mathbf{y}) = R_{N,M}\,.
\end{align}

\paragraph{Backward pass.}
The gradient $E_{i,j} = \partial R_{N,M} / \partial d(x_i, y_j)$ is computed by a reverse-direction Dynamic Programming (DP) over the same $N \times M$ grid, starting from $E_{N,M} = 1$.
This gradient matrix~$E$ is then composed with the distance function's own Jacobian to obtain gradients with respect to the input sequences.

\paragraph{Complexity.}
Both the forward and backward passes require $\mathcal{O}(NM)$ operations per sequence pair and $\mathcal{O}(NM)$ storage for the DP table~$R$.
For a batch of $B$ pairs, the total work is $\mathcal{O}(BNM)$.
Additionally, the pairwise distance matrix~$D$ of shape $(B, N, M)$ must be computed and stored, adding $\mathcal{O}(BNMD)$ computation and $\mathcal{O}(BNM)$ storage.

\section{Limitations of Prior GPU Implementations}
\label{sec:limitations}

Maghoumi's \texttt{pytorch-softdtw-cuda}~\cite{maghoumi2020softdtw} was the first open-source CUDA implementation and pioneered the anti-diagonal parallelization strategy for SoftDTW.
We build on this foundation but identify three limitations it suffers from:

\paragraph{1. Sequence length cap at 1024.}
The implementation assigns one CUDA thread per row of the DP matrix and synchronizes threads within a single block using \texttt{\_\_syncthreads()}.
Since CUDA limits blocks to 1024 threads, sequences longer than 1024 cannot be processed on GPU.
Thus, the code from~\cite{maghoumi2020softdtw} explicitly falls back to CPU for longer sequences.

\paragraph{2. Numerical instability in the backward pass.}
The backward kernel computes transition weights in linear space:
\begin{align}
  a = \exp\!\Big(\frac{R_{i+1,j} - R_{i,j} - d_{i+1,j}}{\gamma}\Big).
  \label{eq:linear_backward}
\end{align}
For a small $\gamma$ (e.g., $\gamma < 0.1$), the exponent can be a large positive number, causing $\exp(\cdot)$ to overflow 
and propagating \texttt{NaN} values through the gradient computation.

\paragraph{3. Full distance tensor materialization.}
The distance matrix $D \in \mathbb{R}^{B \times N \times M}$ is computed upfront via batched matrix multiplication and stored in GPU memory.
For a typical configuration of $B\!=\!32$, $N\!=\!M\!=\!512$, $D_{\mathrm{feat}}\!=\!64$, this tensor alone occupies $32 \times 512 \times 512 \times 4~\text{bytes} \approx 32$\,MB, while the total allocation (including the DP tables, padded copies, and PyTorch autograd overhead) reaches over 8\,GB.

\section{Our Contributions}
\label{sec:contributions}

\subsection{Tiled Anti-Diagonal Execution}
\label{sec:tiling}

Cells on the same anti-diagonal $p = i + j$ are mutually independent: their three predecessors $(i\!-\!1, j\!-\!1)$, $(i\!-\!1, j)$, and $(i, j\!-\!1)$ all lie on diagonals $p\!-\!2$ and $p\!-\!1$, which are already complete. Rather than processing all $N + M - 1$ anti-diagonals within a single kernel launch (which requires $\max(N, M)$ threads per block), we launch a separate kernel for each anti-diagonal, using the host-side sequential ordering of kernel launches as an implicit synchronization barrier (see~\autoref{alg:tiled}).

\begin{algorithm}[h]
\SetAlgoLined
\KwIn{Distance matrix $D$ (or sequences $X, Y$), parameters $\gamma$, bandwidth}
\KwOut{DP matrix $R$, loss $R_{:,N,M}$}
Initialize $R \in \mathbb{R}^{B \times (N+2) \times (M+2)}$ with boundary conditions\;
\For{$p = 0$ \KwTo $N + M - 2$}{
  $i_{\min} \gets \max(0,\, p - M + 1)$\;
  $i_{\max} \gets \min(N - 1,\, p)$\;
  $\ell \gets i_{\max} - i_{\min} + 1$ \tcp*{diagonal length}
  Launch kernel with grid $= (\lceil \ell / 256 \rceil,\; B)$, block $= 256$\;
  \ForEach{\textnormal{thread} $t$ \textnormal{in kernel}}{
    $i \gets i_{\min} + t$, \quad $j \gets p - i$\;
    Compute $R_{i,j}$ via Eq.~\eqref{eq:forward}\;
  }
}
\Return{$R$}\;
\caption{Tiled anti-diagonal forward pass. Each kernel launch processes one anti-diagonal across all batch elements.}
\label{alg:tiled}
\end{algorithm}
Each kernel is launched over a 2D grid of shape $(\lceil \ell_p / 256 \rceil, B)$, where $\ell_p$ is the length of diagonal~$p$.
This allows anti-diagonals of arbitrary length to be tiled across multiple blocks, removing the 1024-thread constraint entirely.
For sequences where $\max(N, M) \leq 1024$, we retain the original single-launch kernel as a fast path to avoid Python-level loop overhead.

\subsection{Log-Space Backward Pass}
\label{sec:logspace}

We reformulate the backward DP entirely in log-space.
Let $\bar{E}_{i,j} = \log E_{i,j}$.
The recurrence becomes
\begin{align}
  \bar{E}_{i,j} = \mathrm{logsumexp}\!\big(\bar{E}_{i+1,j} + \alpha,\;\; \bar{E}_{i,j+1} + \beta,\;\; \bar{E}_{i+1,j+1} + \delta\big)
  \label{eq:log_backward}
\end{align}
where $\alpha = (R_{i+1,j} - R_{i,j} - d_{i+1,j}) / \gamma$ and similarly for $\beta$, $\delta$.
The numerically stable $\mathrm{logsumexp}$ is
\begin{align}
  \mathrm{logsumexp}(a, b, c) = m + \log\!\big(e^{a-m} + e^{b-m} + e^{c-m}\big), \quad m = \max(a, b, c).
\end{align}

The final $\exp(\cdot)$ is applied only once after the full backward DP completes: $E = \exp(\bar{E})$.
This ensures that all intermediate computations remain in log-space, where the max-shift trick keeps values numerically bounded regardless of~$\gamma$.

\subsection{Fused Distance Computation}
\label{sec:fused}

Both unfused and fused modes employ the algebraic identity to compute squared Euclidean distance:
\begin{equation}
  \|x_i - y_j\|^2 = \|x_i\|^2 - 2 \langle x_i, y_j \rangle + \|y_j\|^2,
  \label{eq:sqeuclidean}
\end{equation}
This is substantially more efficient than materializing all $(N \times M)$ pairwise squared differences directly.
Precomputing squared norms (shapes $B \times N$ and $B \times M$) and using batched matrix multiplication for dot products reduces computation compared to naive expansion.

\paragraph{Fused vs. Unfused Trade-offs.}
The key difference lies in when the distance tensor is materialized:

\textit{Unfused (standard) mode:} Precomputes and \textit{stores} the full distance matrix $D \in \mathbb{R}^{B \times N \times M}$ upfront using Eq.~\eqref{eq:sqeuclidean}.
This enables $O(1)$ distance lookups during DP kernel execution, resulting in minimal kernel runtime.
However, the memory footprint is high: the distance tensor accounts for $\approx 32$\,MB to several GB depending on $N$, $M$, and $B$.

\textit{Fused mode:} Eliminates the stored distance tensor by \textit{recomputing} distances on-the-fly within the CUDA kernel, also using Eq.~\eqref{eq:sqeuclidean}.
Each thread computes the cost for its assigned cell by directly reading from the input sequences $X \in \mathbb{R}^{B \times N \times D}$ and $Y \in \mathbb{R}^{B \times M \times D}$.
At each of the $N + M - 1$ anti-diagonal steps, neighboring distances are recomputed in the backward pass.
This trade-off results in a \textit{10--15$\times$ runtime increase} (Table~\ref{tab:runtime}) but a \textit{40--98\% memory reduction} compared to unfused, reducing distance-related storage from $\mathcal{O}(BNM)$ to $\mathcal{O}(B(N+M))$.

For typical research workflows, fused mode is the recommended default when GPU memory is constrained (common with large batch sizes or very long sequences); unfused mode may be preferred when runtime dominates and memory is plentiful.

The backward kernel similarly recomputes the three neighboring costs needed for each cell's gradient update, reading $X$ and $Y$ directly.
The gradient with respect to the inputs is then obtained via efficient marginal reductions:
\begin{align}
  \frac{\partial\, \mathrm{sdtw}}{\partial x_{i}} &= 2 \Big( x_i \sum_j E_{i,j} - \sum_j E_{i,j}\, y_j \Big), \\
  \frac{\partial\, \mathrm{sdtw}}{\partial y_{j}} &= 2 \Big( y_j \sum_i E_{i,j} - \sum_i E_{i,j}\, x_i \Big),
\end{align}
which are computed as matrix multiplications in PyTorch without ever materializing~$D$.

\subsection{Barycenter Computation}
\label{sec:barycenter}
\begin{figure}[ht]
  \centering
  \includegraphics[width=0.99\linewidth]{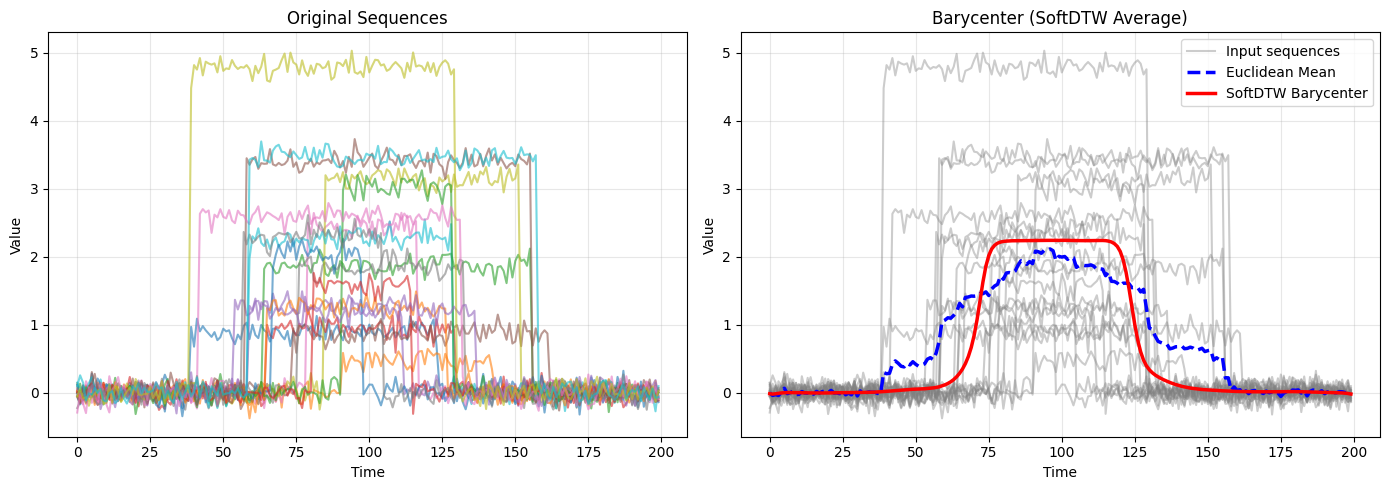}
  \caption{SoftDTW Barycenter on synthetic block-wave data.}
  \label{fig:barycenter}
\end{figure}
We additionally provide a gradient-based SoftDTW barycenter~\cite{cuturi2017soft, petitjean2011global} for computing DTW-space averages of a time-series collection.
The barycenter $\bar{\mathbf{z}}$ minimizes $\sum_{k=1}^{K} \mathrm{sdtw}_\gamma(\bar{\mathbf{z}}, \mathbf{x}^{(k)})$ via Adam optimizer~\cite{kingma2014adam}. 

\section{Benchmarks}
\label{sec:benchmarks}
All benchmarks were run on an NVIDIA GTX 1080 GPU with PyTorch and Numba CUDA.
We report peak GPU memory (forward + backward) in Table~\autoref{tab:memory} and wall-clock runtime in Table~\autoref{tab:runtime}, each averaged over 5 iterations after a warmup pass.
Figure~\autoref{fig:benchmarks} visualizes these results as a function of sequence length ($L$) and feature dimension ($D$).
We compare against Maghoumi's \texttt{pytorch-softdtw-cuda}~\cite{maghoumi2020softdtw}.
Parameters: $\gamma = 1.0$, no bandwidth constraint, squared Euclidean distance.
Note: the benchmark tables and figures use $L$ for sequence length (with $N\!=\!M\!=\!L$); the preceding sections use $N$ and $M$ to denote the (potentially unequal) lengths of the two input sequences.

\begin{table}[ht]
\footnotesize
\setlength{\tabcolsep}{2pt}
\begin{minipage}[t]{0.53\linewidth}
  \centering
  \caption{Peak GPU memory (MB) for forward + backward pass. OOM = out-of-memory.}
  \label{tab:memory}
  \begin{tabular}{@{}rrr rrr r@{}}
  \toprule
  $B$ & $L$ & $D$ & Maghoumi & Ours (unfused) & Ours (fused) & Savings \\
  \midrule
  16 & 128 & 64 & 275 & 26 & 23 & 91.6\% \\
  16 & 512 & 64 & 4{,}136 & 137 & 89 & 97.9\% \\
  16 & 1024 & 64 & OOM & 481 & 289 & --- \\
  16 & 2048 & 64 & OOM & 1{,}842 & 1{,}074 & --- \\
  \midrule
  32 & 128 & 64 & 534 & 35 & 28 & 94.7\% \\
  32 & 512 & 64 & 8{,}256 & 257 & 161 & 98.1\% \\
  32 & 1024 & 64 & OOM & 946 & 562 & --- \\
  32 & 2048 & 64 & OOM & 3{,}672 & 2{,}134 & --- \\
  \bottomrule
  \end{tabular}
  \vspace{0.3em}
\end{minipage}%
\hfill
\begin{minipage}[t]{0.45\linewidth}
  \centering
  \caption{Wall-clock runtime (ms) for forward + backward pass.}
  \label{tab:runtime}
  \begin{tabular}{@{}rrr rrr@{}}
  \toprule
  $B$ & $L$ & $D$ & Maghoumi & Ours (unfused) & Ours (fused) \\
  \midrule
  16 & 128 & 64 & 7.7 & 1.8 & 47.0 \\
  16 & 512 & 64 & 83.2 & 16.0 & 200.8 \\
  16 & 1024 & 64 & OOM & 94.0 & 891.6 \\
  16 & 2048 & 64 & OOM & 685.1 & 5{,}970 \\
  \midrule
  32 & 128 & 64 & 13.4 & 2.4 & 57.9 \\
  32 & 512 & 64 & 2{,}791 & 41.8 & 429.5 \\
  32 & 1024 & 64 & OOM & 189.3 & 2{,}877 \\
  32 & 2048 & 64 & OOM & 1{,}069 & 13{,}319 \\
  \bottomrule
  \end{tabular}
  \vspace{0.3em}
\end{minipage}
\end{table}

\paragraph{Discussion.}
The unfused mode is consistently faster due to more efficient memory access patterns.
The fused mode is 10--15$\times$ slower than unfused but reduces memory by 40\% (unfused~$\to$~fused) and up to 98\% compared to Maghoumi.
This makes fused mode the recommended choice when GPU memory is the bottleneck, a common scenario in training settings with large batch sizes or long sequences.
Importantly, our implementation is the only one that supports $N > 1024$ on GPU at all.
~\autoref{fig:benchmarks} visualizes these trends across sequence lengths and feature dimensions.

\section{Limitations and Future Work}
\label{sec:limitations_future}

\paragraph{Fused mode runtime.}
The fused backward kernel recomputes three distance values per cell per anti-diagonal step, resulting in significant runtime overhead compared to the unfused path.
Shared-memory caching of $X$ and $Y$ tiles could reduce this gap.

\paragraph{Normalization constraint.}
The normalized SoftDTW variant $\mathrm{sdtw}(\mathbf{x}, \mathbf{y}) - \tfrac{1}{2}[\mathrm{sdtw}(\mathbf{x}, \mathbf{x}) + \mathrm{sdtw}(\mathbf{y}, \mathbf{y})]$~\cite{blondel2021differentiable} currently requires $N = M$.
Extending to unequal lengths requires padding or interpolation.

\paragraph{Kernel launch overhead.}
The tiled path issues $N + M - 1$ separate kernel launches from Python.
For very long sequences (e.g., $N = M = 5000$), this amounts to $\sim$10{,}000 launches.
A persistent-kernel or CUDA-graph approach could amortize this overhead.

\paragraph{Mixed precision.}
The current implementation operates in FP32.
FP16 or BF16 support could further reduce memory and improve throughput on modern GPUs with tensor cores.

\section{Conclusion}
\label{sec:conclusion}

We presented \texttt{softdtw-cuda-torch}, a GPU-accelerated SoftDTW library for PyTorch that removes the 1024 sequence-length limit via tiled anti-diagonal execution, ensures numerical stability through a log-space backward pass, and achieves up to 98\% GPU memory reduction via fused on-the-fly distance computation.
The library is open-source under the MIT license and available at \url{https://github.com/BGU-CS-VIL/sdtw-cuda-torch}.

\paragraph{Acknowledgments.}
We thank Mehran Maghoumi for the original \texttt{pytorch-softdtw-cuda} implementation~\cite{maghoumi2020softdtw}, which pioneered GPU-accelerated SoftDTW and served as a foundation for our work.

\bibliographystyle{unsrt}
\bibliography{references}

\end{document}